\documentclass[10pt,twocolumn,letterpaper]{article}

\usepackage[pagenumbers]{cvpr} 

\usepackage{graphicx}
\usepackage{amsmath}
\usepackage{amssymb}
\usepackage{booktabs}
\usepackage{multirow}
\usepackage{adjustbox}
\usepackage[overload]{textcase}
\usepackage{floatrow}
\usepackage{booktabs}
\usepackage{caption}
\usepackage{arydshln}
\usepackage{bbm}

\usepackage{lipsum}
\usepackage{mathtools}
\usepackage{cuted}

\usepackage{floatrow}
\usepackage{algorithm}
\usepackage{algpseudocode}

\usepackage{listings}
\usepackage{xcolor}
\usepackage{color, colortbl}
\definecolor{Gray}{gray}{0.9}

\lstdefinestyle{mystyle}{
    commentstyle=\color{codegreen},
    keywordstyle=\color{magenta},
    numberstyle=\tiny\color{codegray},
    stringstyle=\color{codepurple},
    basicstyle=\ttfamily\footnotesize,
    breakatwhitespace=false,         
    breaklines=true,                 
    captionpos=b,                    
    keepspaces=true,                 
    numbersep=5pt,                  
    showspaces=false,                
    showstringspaces=false,
    showtabs=false,                  
    tabsize=2
}

\usepackage[pagebackref,breaklinks,colorlinks]{hyperref}

\usepackage[capitalize]{cleveref}
\crefname{section}{Sec.}{Secs.}
\Crefname{section}{Section}{Sections}
\Crefname{table}{Table}{Tables}
\crefname{table}{Tab.}{Tabs.}

\def\cvprPaperID{11184} 
\def\confName{CVPR}
\def\confYear{2022}

\begin{document}

\title{Cross-modal Manifold Cutmix for \\ Self-supervised Video Representation Learning}

\author{
  Srijan Das \thanks{work at Stony Brook University.}\\
  UNC Charlotte\\
  {\tt sdas24@uncc.edu}\\
  \and
  Michael Ryoo\\
  Stony Brook University\\
  {\tt mryoo@cs.stonybrook.edu}\\
}

\maketitle
\begin{abstract}
   Contrastive representation learning of videos highly relies on the availability of millions of unlabelled videos. This is practical for videos available on web but acquiring such large scale of videos for real-world applications is very expensive and laborious.  
   Therefore, in this paper we focus on designing video augmentation for self-supervised learning, we first analyze the best strategy to mix videos to create a new augmented video sample. Then, the question remains, \textit{can we make use of the other modalities in videos for data mixing?} To this end, we propose \textbf{Cross-Modal Manifold Cutmix} (CMMC) that inserts a video tesseract into another video tesseract in the feature space across two different modalities. We find that our video mixing strategy \textbf{STC-mix}, i.e. preliminary mixing of videos followed by CMMC across different modalities in a video, improves the quality of learned video representations. We conduct thorough experiments for two downstream tasks: action recognition and video retrieval on two small scale video datasets UCF101, and HMDB51. We also demonstrate the effectiveness of our STC-mix on NTU dataset where domain knowledge is limited.
   We show that the performance of our STC-mix on both the downstream tasks is on par with the other self-supervised approaches while requiring less training data.  
\end{abstract}

\section{Introduction}
\label{sec:intro}
The recent advancements in self-supervised representation is credited to the success of using discriminative contrastive loss such as InfoNCE~\cite{infonce}. Given a data sample, contrastive representation learning focus on discriminating its transformed version from a large pool of other instances or their transformations. Thus, the concept of contrastive learning while applicable to any domains, its effectiveness rely on the domain-specific inductive bias as the transformations are obtained from the same data instance. For images, these transformations are usually standard data augmentation techniques~\cite{simclr} while in videos, data artifacts that arise from temporal segments within the same video clip~\cite{OPN, sst2, sst3, shufflelearn}. Although these methods rely on large scale videos especially available on web, it would be impractical in real-world applications. For example, understanding activities of daily living is crucial for patient monitoring and smarthome application but these boring videos are not easily available on web. Acquiring such sensitive videos is very expensive in terms of both time and cost~\cite{NTU_RGB+D, STA_iccv}. 


Recently, data mixing strategies \cite{zhang2018mixup, cutoff, yun2019cutmix} have emerged as one of the promising data augmentation for supervised learning methods. These mixing strategies when incorporated with contrastive learning, the quality of the learned representation improves drastically as in~\cite{lee2021imix, verma2021dacl, manifold_mixup} while requiring less training data. Such augmentations introduce semantically meaningful variance for better generalization which is crucial for learning self-supervised representations. While these mixing strategies have been impactful for learning image representations, mixing strategies have been very limitedly explored in the video domain. 

Therefore, in this paper, we study the various data mixing strategies for videos, and propose a new approach to overcome their limitation by mixing across modalities. We first investigate and compare the mixing strategies adopted from the the image domain, and we find that mixing videos by performing simple interpolation of two video cuboids (Mixup) is more effective than inserting a video cuboid within another (Cutmix). This is contradictory to the observations made in the image domain. Furthermore, unlike learning image representations~\cite{lee2021imix}, these data mixing strategies are prone to over-fitting when trained for longer, making them limited for videos. 

Motivated by the success of previous self-supervised techniques exploiting multiple modalities to learn discriminative video representation as in~\cite{look_listen, out_of_time, cooperative_learning, objects_that_sounds, owens2018audiovisual, evolving_losses, miech20endtoend}, in this paper, we pose the following question: \textit{can we take advantage of other modalities for mixing videos while learning self-supervised representation?}

Different modalities of a video like RGB, optical flow, etc. have different distributions and thus, mixing them directly in the input space makes the task of discriminating similar instances from the other instances easier limiting the quality of the learned representation.  
To this end, we propose our Cross-Modal Manifold Cutmix (CMMC), that performs data mixing operation `across different modalities' of a video in their hidden intermediate `representations'. Given the video encoders from different modalities pre-trained with contrastive loss in addition to mixup augmentation, CMMC exploits the underlying structure of the data manifold. This is done by performing cutmix operation in the feature space across space, time and channels. 
To the best of our knowledge, this is the first attempt to perform mixing across channels. The channel mixing of the cross-modal feature map enforces the encoder to learn better semantic concepts in the videos. 
Hence, we train video encoders for different modalities in several stages including the use of mixup strategy in videos and our proposed CMMC. We dub this augmentation strategy as \textbf{STC-mix} which \textit{stands for space, time, channel Mix} for contrastive representation learning.

Empirically, we confirm that STC-mix being easy to implement, significantly improves contrastive representation learning for videos. We show that STC-mix can effectively learn self-supervised representation with \textit{small availability of data for pretext task} and can also \textit{take advantage of other modalities of the videos} through manifold mixing strategy. We thoroughly evaluate the quality of the learned representation on two downstream tasks action recognition and retrieval, on UCF101 and HMDB51. We demonstrate the improvement in transferability of the representation learned with STC-mix by conducting training on a large scale dataset Kinetics-400 and then finetuning on smaller datasets. Furthermore, we corroborate the robustness of STC-mix by observing similar improvements on RGB-D videos for the task of action recognition.

\section{Background}\label{bck}
In this section, we first review a general contrastive learning framework used for learning self-supervised video representation. Then, we review a data mixing formulation for self-supervision in the image domain.
Let $\mathcal{X} \in \mathcal{R}^{T \times 3 \times H \times W}$ be a sequence of video. The objective is to learn a mapping $f:\mathcal{X} \rightarrow z$ where $z \in \mathcal{R}^D $, that can be effectively used to discriminate video clips for various downstream tasks, e.g. action recognition , retrieval, etc.

\noindent \textbf{Contrastive Learning.} Assume a set of augmentation transformations $A$ is applied to $\mathcal{X}$. So, for a particular video there exists a positive (say $\widetilde{\mathcal{X}}$) whereas the other transformed videos in a mini-batch are considered as negatives. The encoder $f(.)$ and its exponential average model $\widetilde{f}(.)$ maps the positives and negatives respectively to embedding vectors. Therefore, the contrastive loss for a sample $\mathcal{X}_i$ is formulated as 
\begin{equation}
    \mathcal{L}(\mathcal{X}_i) = -\mathrm{log}\frac{\mathrm{exp}(z_i \cdot \widetilde{z}_i/\tau)}{\sum \limits_{j=0}^{\mathcal{N}}\mathrm{exp}(z_i \cdot \widetilde{z}_j/\tau)}
\end{equation}
where $\tau$ is a scaling temperature parameter and $\mathcal{N}$ is the set of negatives. Note that the embedding vectors $z_i$ and $\widetilde{z}_i$ are L2-normalized before the loss computation. Thus, the loss $\mathcal{L}$ optimizes the video instances such that the representation of the video instances with the same view are pulled towards each other while pushing away from the other instances.

\noindent \textbf{Data Mix for Contrastive Learning.} We revisit the formulation proposed in i-mix~\cite{lee2021imix} for mixing data within a batch for contrastive representation learning. Let $y_i \in \{0, 1\}^{BS}$ be the virtual labels of the input $\mathcal{X}_i$ and $\mathcal{\widetilde{X}}_i$ in a batch, where $y_{i,i} = 1$ and $y_{i,j\neq i} = 0$. Then, the $(\mathcal{N}+1)-$ way discrimination loss for a sample in a batch is:
\begin{equation}
    \mathcal{L}(\mathcal{X}_i, y_i) = - y_{i,b} \cdot \mathrm{log}\frac{\mathrm{exp}(z_i \cdot \widetilde{z}_b/\tau)}{\sum \limits_{j=0}^{\mathcal{N}}\mathrm{exp}(z_i \cdot \widetilde{z}_j/\tau)}
\end{equation}
where $b$ ranges from 0 to $BS$. Thus, the data instances are mixed within a batch for which the loss is defined as:

\begin{equation}
    \begin{multlined}
   \hspace{-1cm} \mathcal{L}_{Mix}((\mathcal{X}_i, y_i), (\mathcal{X}_r, y_r), \lambda) =\\
   \hspace{0.9cm} \mathcal{L}(\mathrm{Mix}(\mathcal{X}_i, \mathcal{X}_r; \lambda), \lambda y_i + (1-\lambda) y_r)
    \label{imix_eq}
    \end{multlined}
\end{equation}
where $\lambda \sim $ Beta($\alpha$, $\alpha$) is a mixing coefficient, $r \sim rand(BS)$, and Mix() is a mixing operator. 
In the following, we will discuss the appropriate mixing operators in the video domain. 

\section{STC-mix}

In this paper, we use the same i-mix formulation (from the above section) for data mixing while learning discriminative self-supervised representation. First, we investigate the best strategies to define the mixing operator for video domain. Furthermore, we introduce a manifold mixing strategy to make use of the other modalities freely available in videos for data mixing. We integrate both these data augmentation strategies, together called STC-mix for contrastive representation learning of videos.

\subsection{Mixing Operator for Videos}\label{mixup}
Unlike mixing operations in images as in ~\cite{zhang2018mixup, manifold_mixup, cutoff, yun2019cutmix}, videos have temporal dimension. 
For the Mixing operation defined in equation~\ref{imix_eq}, it is straightforward to extend the existing image mixing strategies to videos. But, we argue that handling temporal dimension in videos is not equivalent to handling spatial dimension in images. Mixup~\cite{manifold_mixup} in videos perform weighted averaging of two spatio-temporal stack of frames. 
In contrast to cutmix operator~\cite{yun2019cutmix}, mixup operator retains the temporal information in videos and thus facilitates the contrastive representation learning. We empirically corroborate this observation in the experimental analysis.
In addition to this, videos possess different modalities like optical flow that can be computed without any supervision. The question remains that can we make use of other modalities in videos for mixing instances while learning contrastive representation? To this end, we introduce \textbf{Cross-Modal Manifold Cutmix} (CMMC) strategy for mixing video instances across different modalities which is discussed in the next section.

\begin{figure*}
\scalebox{0.7}{
   \includegraphics[width=1\linewidth]{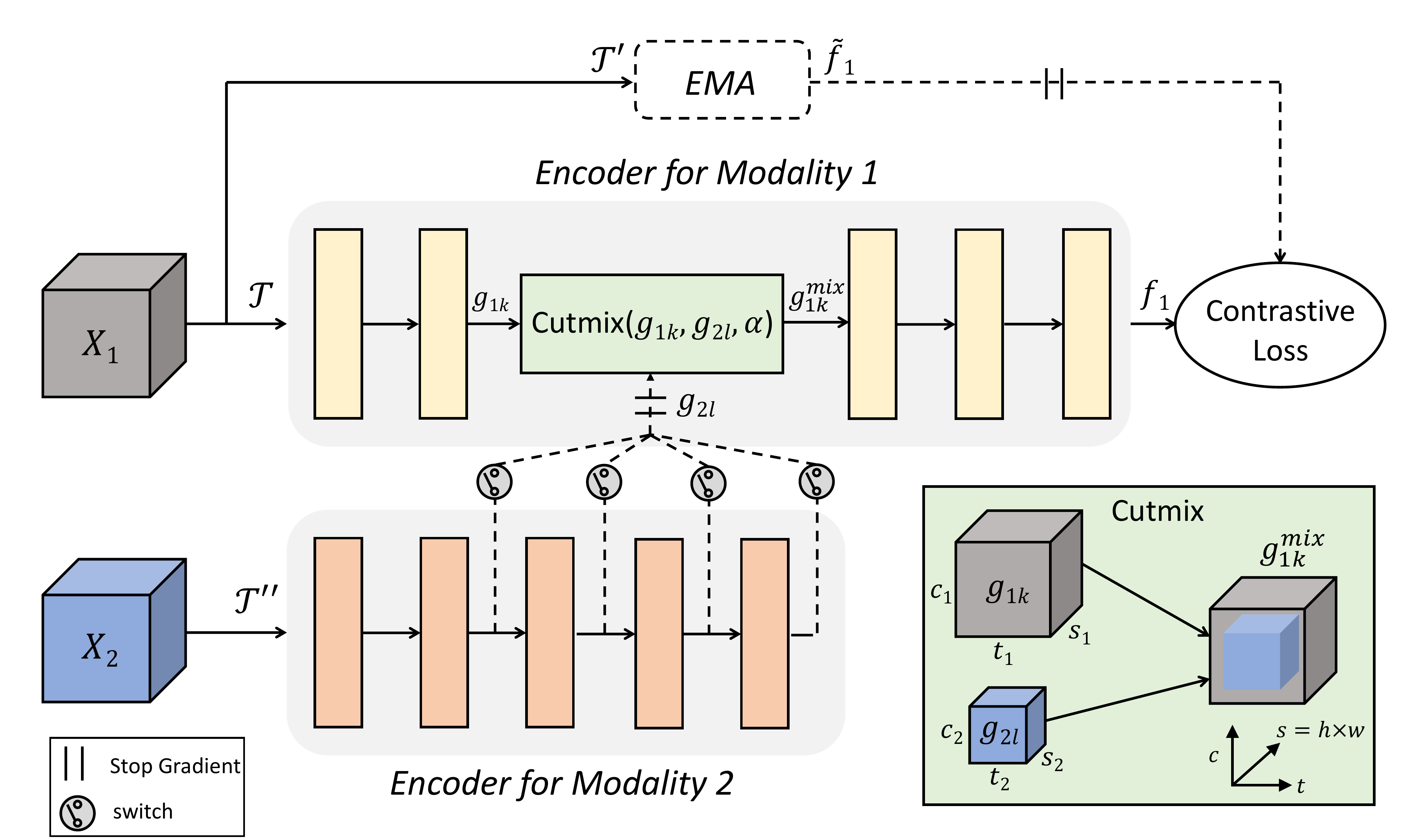} 
   \caption{Cross-Modal Manifold Cutmix (CMMC) trains a visual encoder $f_1$ for Modality 1 with $\mathcal{X}_1$ as input while mixing with input $\mathcal{X}_2$ from modality 2. The mixing operation between the cross-modal feature maps $g_{1k}$ and $g_{2l}$ is Cutmix which is illustrated elaborately at the right. For brevity, we set $k=2$. Thus, $l$ is randomly selected from the set \{2,3,4,5\} through a switch. }
\label{framework}} 
\end{figure*}

\subsection{Cross-Modal Manifold Cutmix}
Different modalities in videos is an additional information that are often exploited for self-supervised learning as in~\cite{coclr, li2021crossclr}. In contrast to these approaches, we simply propose to mix these different modalities as another data augmentation strategy for self-supervised representation. However, the dissimilarity in distribution between the different modalities (say, RGB and optical flow) in videos makes it harder to mix them at input space. Consequently, we propose Cross-Modal Manifold Cutmix to mix such cross-modal representations in the hidden representation space.  

As an extension of the previous notation, we now consider two different modalities $\mathcal{X}_{1i}$ and $\mathcal{X}_{2i}$ for a given video clip $\mathcal{X}_i$. The objective of the self-supervised task is to learn discriminative video representation, i.e. to learn functions $f_1(\cdot)$ and $f_2(\cdot)$. We decompose the encoder function by $f_1(\mathcal{X}_{1i}) = f_{1k}(g_{1k}(\mathcal{X}_{1i}))$, where $g_{1k}$ is a part of the video encoder for modality 1 with $k$ layers that maps the input data $\mathcal{X}_{1i}$ to a hidden representation. Similarly, $f_{1k}$ maps the hidden representation $g_{1k}(\mathcal{X}_{1i})$ to the embedding vector $z_{1i}$. Note that we already have trained video encoders $f_{1i}(\cdot)$ and $f_{2i}(\cdot)$ by exploiting the above mentioned Mixup strategy among the video instances in a mini-batch while optimizing the contrastive loss.
Now, CMMC is trained in a 4 stage fashion. In the first stage, we train the encoder $f_{1i}(\cdot)$ of modality 1 in 5 steps as illustrated in figure~\ref{framework}. First, we select random layers $k$ and $l$ from a set of eligible layers in $f_{1i}(\cdot)$ and $f_{2i}(\cdot)$ respectively such that $k \leq l$. This set excludes the input space. Second, we fed a pair of input $\mathcal{X}_{1i}$ and $\mathcal{X}_{2r}$ to their respective video encoders $f_1$ and $f_2$ until they reach layer $k$ and layer $l$ respectively. We obtain $g_{1k}(\mathcal{X}_{1i})$ and $g_{2l}(\mathcal{X}_{2r})$ - a hidden representation  (spatio-temporal tesseract) of both videos in modality 1 and 2. Third, we perform a data mixing among the hidden representations across two modalities as:
\begin{equation}
    g_{1k}^{mix}, \lambda = \mathrm{CutMix}(g_{1k}, g_{2l}; \alpha)
\end{equation}
\begin{equation}
    y_{1k}^{mix} = \lambda y_{1i} + (1-\lambda) y_{2r} \label{mixlabel}
\end{equation}
where ($y_{1i}$, $y_{2r}$) are one-hot labels, hyper-parameter $\alpha = 1$, and the mixing operator is cutmix as in~\cite{yun2019cutmix} which returns the mixing coefficient $\lambda$ along with the mixed data. For brevity, we omit the input instances in the equation. Fourth, we continue the forward pass in $f_1(\cdot)$ only from layer $k$ to the output embedding, now we denote by $z_{1i}^{mix}$. Fifth, this embedding is used to compute the ($\mathcal{N}+1$)-way discrimination loss which is reformulated as:
\begin{equation}
    \mathcal{L}(\mathcal{X}_{1i}, y_{1i}) = - y_{1i,b}^{mix} \cdot \mathrm{log}\frac{\mathrm{exp}(z_{1i}^{mix} \cdot \widetilde{z}_{1b}/\tau)}{\sum \limits_{j=0}^{\mathcal{N}}\mathrm{exp}(z_{1i}^{mix} \cdot \widetilde{z}_{1j}/\tau)}
\end{equation}
The computed gradients are backpropagated through the entire video encoder $f_1(\cdot)$ of modality 1 only. It is to be noted that the video encoder $f_2(\cdot)$ of modality 2 is not trained in this stage.
In the second stage, we train the video encoder $f_2(\cdot)$ for modality 2 while freezing the updated learned weights of $f_1(\cdot)$. We continue this cycle twice for each modality, and hence 4 stages to learn the self-supervised video representation in $f_1(\cdot)$ and $f_2(\cdot)$ . We provide the Pseudocode  of one stage of CMMC for training encoder $f_1(\cdot)$ in Algorithm~\ref{algo}. 

\begin{algorithm}
\caption{Pseudocode of One stage CMMC}
\label{algo}
\begin{algorithmic}[1]
\State $\alpha, k \gets 1.0, \text{rand}(1, N)$ \Comment{$N$ is the layers in the encoder}
\State $l \gets \text{rand}(k, N)$
\State $\mathbf{x}_1, \widetilde{\mathbf{x}}_1 \gets \mathcal{T}(\mathbf{X}_1)$ \Comment{Two views of the modality 1}
\State $\mathbf{x}_2 \gets \mathcal{T}''(\mathbf{X}_2)$ \Comment{modality 2 data}
\State $\mathbf{g}_{1k} \gets f_1.\text{partial\_forward}(\mathbf{x}_1, 0, k)$
\State $\mathbf{g}_{2l} \gets f_2.\text{partial\_forward}(\mathbf{x}_2, 0, l)$
\State $\mathbf{g}_{1k}^{mix}, \text{labels\_new}, \lambda \gets \text{CutMix}(\mathbf{g}_{1k}, \mathbf{g}_{2l}, \alpha)$
\State $\mathbf{z}_1 \gets \text{normalize}(f_1.\text{partial\_forward}(\mathbf{g}_{1k}^{mix}, l, N))$
\State $\mathbf{z}_2 \gets \text{normalize}(\widetilde{f}_1.\text{forward}(\widetilde{\mathbf{x}}_1))$
\State $\mathbf{z}_2, \mathbf{g}_{2l} \gets \mathbf{z}_2.\text{detach}(), \mathbf{g}_{2l}.\text{detach}()$ \Comment{no gradient flow}
\State $\text{labels} \gets \text{zeros}(len(\mathbf{x}_1))$
\State $\text{logits} \gets \text{matmul}(\mathbf{z}_1, \mathbf{z}_2^T) / t$ \Comment{$t$ is the temperature}
\State $\text{loss} \gets \lambda \times \text{CE}(\text{logits}, \text{labels}) + (1 - \lambda) \times \text{CE}(\text{logits}, \text{labels\_new})$
\end{algorithmic}
\end{algorithm}

Thus, to sum up STC-mix consists of initially training video encoders of modality 1 and modality 2 independently with infoNCE loss as in~\cite{simclr} and applying mixup augmentation. Then, we perform CMMC among the hidden representations of data from modality 1 and 2 in 4 stages. This is performed by alternation training strategy as in~\cite{coclr} to make use of the latest learned representations in the cross-modal network. The final learned model is obtained after two cycles of training encoder of each modality. 

\textbf{CutMix in feature space.} Here, we explain how the cutmix operator is applied on the video tesseracts in the feature space. Assume that the hidden representation of the input video sequence $\mathcal{X}_{1i}$ in modality 1, $g_{1k} \in \mathcal{R}^{c_1 \times t_1 \times h_1 \times w_1}$, where $c_1$ represents channel, $t_1$ time, and $s_1 = h_1 \times w_1$ is the spatial resolution. We generate a new representation $g_1^{mix}$ by combining the hidden representations $g_{1k}$ and $g_{2l}$. These hidden representations $g_{1k}$ and $g_{2l}$ may differ such that $(t_1, s_1) \geq (t_2, s_2)$ and $c_1 \leq c_2$. Therefore, we define a cutmix operation that combines the video tesseracts in space, time and across channels. We define the combining operation as 
\begin{equation}
    g_1^{mix} = M \odot g_{1k} + (1 - M) \odot g_{2l}
\end{equation}
where $M \in \{0,1\}^{c_1 \times t_1 \times h_1 \times w_1}$ is a binary tensor mask which is decided by sampling the bounding box coordinates $bbox = (b_{c1}, b_{c2}, b_{t1}, b_{t2}, b_{h1}, b_{h2}, b_{w1}, b_{w2})$ from a uniform distribution. In order to preserve the temporal information in a video, we fix $(b_{t1}, b_{t2}) = (0, t_2)$. Similarly, we preserve the channel information processed by the video encoder $f_2(\cdot)$ by fixing  $(b_{c1}, b_{c2}) = (0, c_2)$. 
Thus, the bounding box selection follows a random sampling of a center coordinate $(b_{wc}, b_{hc})$ from $(U(0, w_2), U(0, h_2))$. The corner points of the bounding box are determined by
where $\lambda \sim U(0,1)$. Even by fixing $(b_{t1}, b_{t2})$ and $(b_{c1}, b_{c2})$, the resultant video tesseract in one modality may not match the dimension of the video tesseracts in other modality across channel and time, if $k < l$. 
So, we select a 4D bounding box with coordinates $(M_{c1}, M_{c2}, M_{t1}, M_{t2}, M_{h1}, M_{h2}, M_{w1}, M_{w2})$ within the defined binary mask $M$. 
We randomly sample a center coordinate $(M_{cc}, M_{tc}, M_{hc}, M_{wc})$ from $(U(0, c_1), U(0, t_1))$, $(U(0, h_1)$, and $U(0, w_1))$ respectively. The end points of the binary mask $M$ are determined by
\begin{equation}
\begin{array}{l}
    M_{c1}, M_{c2} = M_{cc} - \frac{c2}{2}, M_{cc} + \frac{c2}{2} \\
    M_{t1}, M_{t2} = M_{tc} - \frac{t2}{2}, M_{tc} + \frac{t2}{2} \\
    M_{h1}, M_{h2} = M_{hc} - \frac{h2}{2}, M_{hc} + \frac{h2}{2} \\
    M_{w1}, M_{w2} = M_{wc} - \frac{w2}{2}, M_{wc} + \frac{w2}{2}
    \end{array}
\end{equation}
For the region within this bounding box, the values in the binary mask is filled with 0, otherwise 1. A new mixing coefficient is computed by $1 - \lambda_{new} = \sum_{c,t,w,h} M_{c,t,w,h}$ denoting the complementary of the proportion of volume occupied by $M$. This new mixing coefficient $\lambda_{new}$ is returned by the cutmix function to compute the mixed labels in equation~\ref{mixlabel}.

Thus, we perform a mix operation in videos across all the dimensions including spatial, temporal and channels. However, we preserve the temporal properties of the video instances by retaining a proportion of channel information. This makes cutmix operation effective in the feature space.  

\section{Experiments}
In this section, we describe the datasets used in our experimental analysis, implementation details, and evaluation setup. We present ablation studies to illustrate the effectiveness of STC-mix video data augmentation and also, provide an exhaustive state-of-the-art comparison with our STC-mix models. 
\subsection{Datasets}
We use two video action recognition datasets: UCF101~\cite{ucf} and Kinetics-400~\cite{kinetics} for self-supervised training of the video encoders. UCF101 contains 13k videos with 101 human actions and Kinetics-400 (K400) contains 240k video clips with 400 human actions.  
To show the robustness of STC-mix on domains with limited data and other modalities, we also use an RGB+D action recognition dataset: NTU-RGB+D~\cite{NTU_RGB+D} for self-supervised training of a video and skeleton encoder. NTU-RGB+D (NTU-60) contains 58k videos with 60 human action, all performed indoors. Note that we use the videos or skeleton sequences from the training set only for the self-supervised pre-training.
Downstream tasks are evaluated on split1 of UCF101 and split1 of HMDB51~\cite{kuehne2011hmdb}, which contains 7k videos with 51 human actions. For evaluation on NTU-60, we evaluate on the validation set of NTU-60 on Cross-subject (CS) and Cross-View (CV) protocols.  
\subsection{Implementation details}\label{implementation}
STC-mix is a simple data augmentation strategy that requires cutmix operation in the feature space which is adopted from~\cite{yun2019cutmix} followed by our temporal and channel mixing. The input modalities in our experiments consists of RGB, optical flow and skeletons (3D Poses). The optical flow is computed with the un-supervised TV-L1 algorithm~\cite{TV_L1} and the same pre-processing procedure is used as in~\cite{i3d}. For the skeleton experiments, the skeleton data $\mathcal{X} \in \mathcal{R}^{C \times T \times V}$ is acquired using KinectV2 sensors, where coordinate feature $C=3$, \textit{\#} joints $V=25$, and \textit{\#} frames $T=50$. Following the pre-processing steps in~\cite{li2021crossclr}, we compute the joints and motion cues. 
For all the RGB and optical flow models, we choose S3D~\cite{s3d} architecture as the backbone whereas for the skeleton model, we choose ST-GCN~\cite{stgcn2018aaai} with channels in each layer reduced by $1/4$ times as the backbone.
For self-supervised representation learning, we adopt a momentum-updated history queue to cache a large number of video features as in MoCo~\cite{he2019moco}. We attach a non-linear projection head, and remove it for downstream task evaluations as done in SimCLR~\cite{simclr}.

For our experiments with RGB and optical flow, we use 32 $128 \times 128$ frames of RGB (or flow) input, at 30 fps. For additional data augmentation, we apply clip-wise consistent random crops, horizontal flips, Gaussian blur and color jittering. We also apply random temporal cropping from the same video as used in~\cite{coclr}. For training a MoCo model with STC-mix data augmentation, we initially train the RGB and Flow networks for 300 epochs with mixup data augmentation independently. The mixup operation is applied in the input space. Then, we train these pre-train networks with CMMC in 4 stages. In each stage, a network with one input modality is trained for 100 epochs by freezing the network with other modality. In the next stage, we reverse the cross-modal networks and continue training the network with other modality. Finally, after the 4 stages, the resultant models are hence trained for 500 epochs in total. For optimization, we use Adam with $10^{-3}$ learning rate and $10^{-5}$ weight decay. All the experiments are trained on 4 and 2 V100 GPUs for K400 and others respectively, with a batch size of 32 videos per GPU.

For our experiments with skeleton sequence, we choose Shear with shearing amplitude $0.5$ and Crop with a padding ratio of $0.6$ as the augmentation strategy as used in~\cite{li2021crossclr}. Note that, for CMMC on skeleton data, we perform cutmix operation only on skeleton vertices followed by channel and temporal mixing. 
For training with CMMC, the GCN encoders are initially trained for 150 epochs on Joint and Motion cues. This is followed by 2 stage training each with 150 epochs, where encoder with one modality is trained and the other is frozen. For optimization, we use SGD with momentum (0.9) and weight
decay (0.0001). The model is trained on 1 V100 with a batch size of 128 skeleton sequences.

\subsection{Evaluation setup for downstream tasks}
For experiments with RGB and optical flow, we evaluate on two downstream tasks: (i) \textbf{action classification} and (ii) \textbf{retrieval}. For action classification, we evaluate on (1) \textbf{linear probe} where the entire encoder is frozen and a single linear layer followed by a \textit{softmax} layer is trained with cross-entropy loss, and (2) \textbf{finetune} where the entire encoder along with a linear and \textit{softmax} layer is trained with cross-entropy loss. Note that the encoders are initialized with the STC-mix learned weights. More details for training the downstream action classification framework is provided in the Apendix.   
For action retrieval, the extracted features from the encoder pre-trained with STC-mix are used for nearest-neighbor (NN) retrieval. We report Recall at $k$ ($R@k$) which implies, if the top $k$ nearest neighbours comprise one video pertaining to the same class, a correct retrieval is counted.

\subsection{Ablation studies on STC-mix}
In this section, we empirically show the effectiveness of our data augmentation strategy for videos. We also investigate the potential causes behind the significant improvement of performance with STC-mix by conducting relevant experiments. 

\begin{table}
\scalebox{0.8}{
  \begin{tabular}{c|c|c|c|c}
\hline
 \multirow{2}{*}{ \textbf{Mixing}}     & \multicolumn{2}{c|}{\textbf{Action cls.}} & \multicolumn{2}{c} {\textbf{Retrieval}} \\ \cline{2-5}
          & \multicolumn{2}{c|} {Linear probe}  & \multicolumn{2}{c} {R@1} \\
\cline{2-5}
 \textbf{Strategies} & \textbf{UCF} & \textbf{HMDB} & \textbf{UCF} & \textbf{HMDB} \\
 \cline{1-5}
MoCo &  38.2 & 15.3 & 29.1 & 9.8 \\
+ \textbf{Mixup}  &  \textbf{49.6} & \textbf{24.9} & \textbf{36.2} & \textbf{16.2} \\
+ T cutmix & 41.7 & 17.3 & 30.2 & 12.9 \\
+ ST cutmix &  19.8 & 14.3 & 32.6& 14.3 \\
+ VideoMix &  45.5 & 21.2 & 34.5 & 15.2 \\
\hline
\end{tabular}
\caption{Different Video mixing strategies are evaluated on downstream action classification and retrieval tasks. All the models are trained on training samples of UCF101 for 200 epochs with RGB input and tested on the validation set of UCF101 and HMDB51.}\label{mixup_table}  }
\end{table}

\noindent \textbf{Which mixing strategy is the best for uni-modal video understanding?} In Table~\ref{mixup_table}, we investigate different video mixing strategies based on mixup and cutmix operator for downstream action classification and retrieval tasks.
For the augmentations based on Cutmix~\cite{yun2019cutmix}, we randomly select a sub-cuboid and plug it into another video. We also consider Videomix ~\cite{yun2020videomix} that performs a cutmix operation across all the frames clipwise consistent. For the virtual labels, we perform label smoothing as defined in Equation~\ref{imix_eq}.
In image domain, cutmix outperforms the mixup strategy in supervised settings~\cite{yun2019cutmix}. However, we find that all strategies using cutmix in temporal dimension (T cutmix), spatio-temporal dimension (ST cutmix) and spatial dimension (VideoMix) performs worse than simple Mixup strategy.
This is because cutmix operation destroys the temporal structure of the videos which is crucial for understanding actions in videos. Similarly, VideoMix where cutmix is performed spatially and not temporally, introduces new contextual information in videos in arbitrary spatial locations. This not only hampers the motion patterns present in the original video but also weakens the similarity between the positive samples in the contrastive loss. Thus, video mixing operators must ensure retention of temporal characteristics in videos.
 \begin{figure}
\centering
\scalebox{0.9}{
\includegraphics[width=0.9\linewidth]{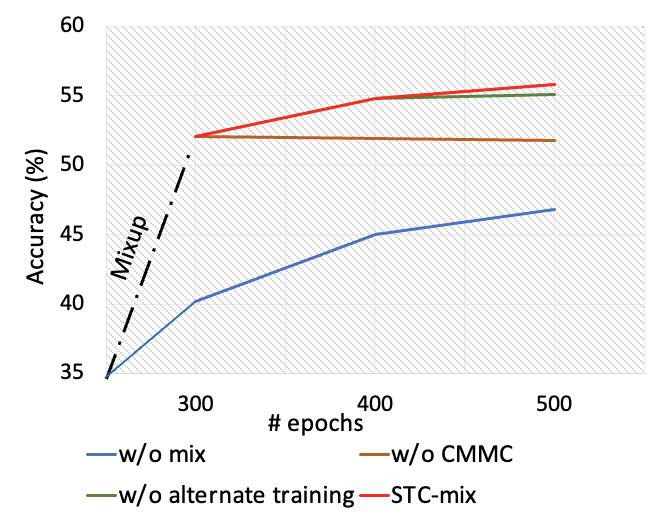}
\caption{Accuracy graph illustrating the improvements in STC-mix with CMMC compared to baselines without using mix, manifold mix, and alternate training in CMMC.}
\label{graph}}
\end{figure}

\noindent \textbf{Why do we need multi-modal mixing strategy for videos?} In Fig.~\ref{graph}, we illustrate the downstream action classification accuracy vs \# epochs plot. We observe that the performance of the model without mix saturates after 500 epochs.
This plot clearly shows the importance of applying data mixing augmentation.
However, the model trained without multi-modal mixing strategy (CMMC) overfits after 300 epochs, whereas the STC-mix models training with multi-modal mixing are still learning discriminative representation. We find that the mixing strategy on video representation learning induces faster training and with CMMC, the models learn cross-modal knowledge without using complicated knowledge distillation techniques as in~\cite{mars,garcia_cross_modal_distillation}. It is to be noted that STC-mix is more beneficial when the cross-modal encoders are trained in alternation strategy. In Fig.~\ref{graph}, we show that the RGB encoder with alternate training outperforms the RGB encoder which is trained for 200 epochs straightaway with the outdated optical flow encoder. The alternate training strategy takes benefit of the most updated cross-modal encoder for data mixing and hence learning more discriminative representation.

\begin{table}
\caption{Comparison of different Cross-Modal manifold mixing strategies. Rand. mix layer indicates if the feature map from the other modality is obtained from the same layer as the primary modality or not. $s=h \times w$, $t$, and $c$ represents spatial, temporal and channel mixing of the feature maps. All the video mix models are trained on training samples of UCF101 for 300 epochs whereas the MoCo model is trained for 500 epochs. $\mathcal{M}$ indicates use of cross-modal information.}
  \centering{%
\scalebox{0.62}{
\begin{tabular}{|c|c|c|c|c|c|c|c|c|c|c|}
\hline
 &\textbf{\small{Cross-Modal}} &    & \textbf{\small{Rand.}} & &  &  & \multicolumn{2}{c|}{\textbf{Action cls.}}  & \multicolumn{2}{c|} {\textbf{Retrieval}} \\  \cline{8-11}
  & \textbf{\small{Manifold Mixing}} & $\mathcal{M}$ & \textbf{\small{mix}} & $s$ & $t$ & $c$     & \multicolumn{2}{c|} {Linear probe}  & \multicolumn{2}{c|} {R@1} \\
\cline{8-11}
& \textbf{\small{strategies}} & & \textbf{\small{layer}} & &  &   & \textbf{UCF} & \textbf{HMDB} & \textbf{UCF} & \textbf{HMDB} \\
 \hline
{\multirow{7}{*}{\rotatebox{90}{\textbf{RGB}}}} &  MoCo (Baseline) & $\times$ & $\times$ & $\times$&$\times$&$\times$& 46.8 & 23.1 & 33.1 & 15.2 \\ 
&  + mixup & $\times$ & $\times$ & \checkmark &\checkmark & \checkmark & 52.8 & 24.4 & 37.6 & 17.6    \\
& + CM mixup & \checkmark & $\times$ & \checkmark &\checkmark & \checkmark & 53.9 & 25.1 & 40.3 & 17.6  \\
 \cline{2-11}
 & \multirow{4}{*} {+ CM cutmix} & \checkmark & $\times$ & \checkmark & $\times$ & $\times$ & 54.6 & 25.5 & 40.4 & 17.8 \\
&& \checkmark&$\times$ & \checkmark & \checkmark & $\times$ &  55.1 & 27.1 & 40.6 & 16.5\\
&&\checkmark &$\times$ & \checkmark & \checkmark & \checkmark &  55.2 & 27.8 & 41.6 & 18.6 \\
& &\checkmark&\checkmark & \checkmark&\checkmark& \checkmark &  \textbf{55.8} & \textbf{28.3} & \textbf{42.8} & \textbf{19.1} \\
\hline
{\multirow{7}{*}{\rotatebox{90}{\textbf{Optical Flow}}}} & MoCo (Baseline) &$\times$& $\times$ & $\times$&$\times$&$\times$& 66.8 & 30.3 & 45.2 & 20.8 \\
&+ mixup & $\times$& $\times$ & \checkmark&\checkmark& \checkmark & 68.6 & 33.1 & 48.7 & 19.5  \\
&+ CM mixup & \checkmark & $\times$ & \checkmark&\checkmark& \checkmark & 70.4 & 33.1 & 51.2 & 21.0 \\
 \cline{2-11}
 & \multirow{4}{*} {+ CM cutmix} &\checkmark & $\times$ & \checkmark & $\times$ & $\times$ & 70.4 & 33.3 & 51.4 & 21.0 \\
& &\checkmark&$\times$ & \checkmark & \checkmark & $\times$ & 71.8  & 34.7 & 52.7 & \textbf{23.1} \\
& &\checkmark&$\times$ & \checkmark& \checkmark & \checkmark & 71.5  & 33.9 & 53.1 & 21.5  \\
&& \checkmark&\checkmark & \checkmark&\checkmark& \checkmark &  \textbf{72.4} & \textbf{34.9} & \textbf{53.9} & \textbf{23.1}  \\
\hline
{\multirow{7}{*}{\rotatebox{90}{\textbf{Two-stream}}}} & MoCo (Baseline) &$\times$& $\times$ & $\times$&$\times$&$\times$ & 68.1 & 33.1 & 49.8 & 21.9 \\
&+ mixup & $\times$ & $\times$ & \checkmark&\checkmark& \checkmark & 71.3  & 36.3 & 53.8 & 24.5 \\
&+ CM mixup & \checkmark & $\times$ & \checkmark&\checkmark& \checkmark & 72.2 & 35.9 & 56.1 & 25.3 \\
 \cline{2-11}
 & \multirow{4}{*} {+ CM cutmix} &\checkmark & $\times$ & \checkmark & $\times$ & $\times$ & 72.1 & 35.8 & 56.9 & 25.1 \\
&& \checkmark&$\times$ & \checkmark & \checkmark & $\times$ & 73.1 & 37.2 & 56.9 & \textbf{27.3} \\
&& \checkmark&$\times$ & \checkmark& \checkmark & \checkmark & 73.4 & 37.2 & 56.4 & 26.5 \\
&&\checkmark &\checkmark & \checkmark&\checkmark& \checkmark & \textbf{74.0} & \textbf{38.1} & \textbf{58.1} & 27.1 \\ 
\hline
\end{tabular}}
}
\label{cutmix_ablation}
\end{table}

\begin{table*}
  \centering
{%
\scalebox{0.75}{
\caption{Unsupervised methods pre-trained on UCF-101 and evaluated on linear probe and fine-tuning for downstream action classification on UCF101 and HMDB51 for different combination of modalities - video ($V$) and flow ($F$). $\dagger$ indicates that the method is reproduced on our configuration.}
\begin{tabular}{c|ccccc|cc|cc}
\hline
\multirow{2}{*}{\textbf{Method}}    & \multirow{2}{*}{\textbf{Network}} & \multirow{2}{*}{\textbf{Res.}} & \multirow{2}{*}{{$\mathcal{D}$}} & \multirow{2}{*}{\textbf{GFLOPs}}& \multirow{2}{*}{{$\mathcal{M}$}}  & \multicolumn{2}{c|} {\textbf{Linear Probe}} & \multicolumn{2}{c} {\textbf{Fine-tune}} \\ 
 &  &  &  &  &  & \textbf{UCF} & \textbf{HMDB} & \textbf{UCF} & \textbf{HMDB} \\
\hline
OPN~\cite{OPN} & VGG & 227 & 14 & 16 & V  & -& -& 59.6 & 23.8 \\
VCOP~\cite{vcop}     & R(2+1) & 112 & 26 & 12.5 & V  & - & - & 72.4 & 30.9 \\
CoCLR-RGB~\cite{coclr}  & S3D & 128 & 23 & 11 & V  & 70.2 & 39.1 & 81.4 & 52.1 \\
$\rho$BYOL$\dagger$~\cite{video_byol}  & $2\times$S3D & 128 & 23 & 22 & V+F  & 70.2 & 37.8 & 84.9 & 57.6 \\
CoCLR~\cite{coclr}  & $2\times$S3D & 128 & 23 & 22 & V+F  & 72.1 & 40.2 & 87.3 & 58.7 \\
\hline
\textbf{STC-mix}  & $2\times$S3D & 128 & 23 & 22 & V+F  & 74.0  & 38.1 & 87.5 & 59.1 \\
\textbf{CoCLR + STC-mix (RGB)}  & S3D & 128 & 23 & 11 & V  & 71.3  & 39.4 & 82.5 & 53.2 \\
\textbf{CoCLR + STC-mix}  & $2\times$S3D & 128 & 23 & 22 & V+F & \textbf{74.7}  & \textbf{40.8} & \textbf{87.9}  & \textbf{59.0} \\
\hline
\end{tabular}\label{SOTA_linear}} 
}\vspace{-0.1in}
\end{table*}

\noindent \textbf{Diagnosis of CMMC.} In Table~\ref{cutmix_ablation}, we provide the results for different configurations of data mixing in the feature space. The objective is to understand the strategies responsible for boosting the performance of the models on UCF101 and HMDB51 for downstream tasks. All the models are initialized with weights obtained from pre-training with mixup and trained for 300 epochs. Note that the initial mixup model is trained for 200 epochs.
First, we show that cross-modal mixup (indicated by + mixup) in the feature space exploiting the cross-modal representation outperforms the traditional manifold mixup (indicated by + mixup) in the feature space~\cite{manifold_mixup} on UFC-101 which does not make use of the cross-modal representation. However, we observe that the action classification accuracy on HMDB51 using optical flow is equivalent for both strategies with or without using cross-modality. This is because HMDB51 mostly consists of static actions with improminent motion patterns which limits the optical flow model to learn motion dominated representation. As a result, this also affects the action classification accuracy on HMDB51 when evaluated with both the streams.
\noindent Next, we show the influence of mixing different dimensions in the hidden representation of a video. We perform experiments with cross-modal cutmix occurring in the spatial dimension ($s$), spatial and temporal ($t$) dimensions, and finally all the dimensions including channels ($c$). Note that the manifold cutmix operation is performed within the same data manifold, i.e. the cutmix is performed across the same random layer between RGB and Flow networks. In contradiction to our previous observation of video mixing in the input space, here the temporal cutmix provides a minor boost to the performance of the downstream tasks. This is supported by the fact that hidden representations of a video retains temporal information due to the preceding convolutional operations on the input sample.  
Finally, we introduce more randomness in CMMC by randomizing the selection of the cross-modal network layer where the mixing takes place. This enables CMMC to take advantage of the features from later layers of the cross-modal network. Thanks to the channel mixing that enables the features from different layers to mix with each other. It also ensures that the lost temporal information is preserved through channels that retains in the original feature map.

\subsection{Comparison to the state-of-the-art}
In this section, we compare STC-mix with the state-of-the-art (SOTA) self-supervised learning (SSL) approaches for video/skeleton action classification and video action retrieval tasks. 

\noindent \textbf{SSL with small training samples.} In Table~\ref{SOTA_linear}, we provide the action classification results on UCF101 and HMDB51 for linear-probing and full finetuning of video encoders with models trained on UCF101. \textit{In this work, we focus on improving the performance of downstream task with UCF101 pre-training due its small size.}
We compare STC-mix with $\rho$BYOL~\cite{video_byol} which is a SOTA SSL method for video representation. 
For a fair comparison, we provide the results of BYOL with both modalities (RGB + Flow). However, it highly relies on the availability large scale unlabelled data in contrast to STC-mix which learns discriminative representation even with less training data.
Furthermore, STC-mix only with its strong data augmentation (mixup + CMMC) outperforms CoCLR by $1.9$\% on UCF101. 
However, the lower action classification accuracy of STC-mix on HMDB51 compared to CoCLR indicates the requirement of positive mining of data samples in contrastive learning as performed in CoCLR. 
It is to be noted that STC-mix is a data augmentation that can be incorporated with any existing self-supervised representation learning methods including CoCLR. Consequently, we have performed STC-mix augmentation while pre-training with CoCLR on UCF-101 in Table~\ref{SOTA_linear} (last two rows). We observe that our proposed augmentation boosts the performance of CoCLR model significantly when finetuned on UCF101 and HMDB51.

\begin{table*}
\centering
{%
\scalebox{0.7}{
\caption{Nearest-Neighbour video retrieval on UCF101 and HMDB51. Testing set clips are used to retrieve training set videos and R@$k$ is reported for $k \in \{1, 5, 10, 20\}$.}
\begin{tabular}{cc|cccc|cccc}
\hline
\multirow{2}{*}{\textbf{Method}}  & \multirow{2}{*}{\textbf{Dataset}} &  \multicolumn{4}{c|}{\textbf{UCF101}} &  \multicolumn{4}{c}{\textbf{HMB51}} \\ 
\cline{3-10}
 & & R@1 & R@5 & R@10 & R@20 & R@1 & R@5 & R@10 & R@20 \\
 \hline
 Jigsaw~\cite{jigsaw} & UCF & 19.7 & 28.5 & 33.5 & 40.0 & - & - & - & - \\
 OPN~\cite{OPN} & UCF & 19.9 & 28.7 & 34.0 & 40.6 & - & - & - & - \\
 RL-method~\cite{buchler} & UCF & 25.7 & 36.2 & 42.2 & 49.2 & - & - & - & - \\
 VCOP~\cite{vcop} & UCF & 14.1 & 30.3 & 40.4 & 51.1 & 7.6 & 22.9 & 34.4 & 48.8 \\
 VCP~\cite{vcp} & UCF & 18.6 & 33.6 & 42.5 & 53.5 & 7.6 & 24.4 & 36.3 & 53.6 \\
 MemDPC~\cite{dpc} & UCF & 20.2 & 40.4 & 52.4 & 64.7 & 7.7 & 25.7 & 40.6 & 57.7 \\
 SpeedNet~\cite{speednet} & K400 & 13.0 & 28.1 & 37.5 & 49.5 & - & - & - & - \\
 CoCLR~\cite{coclr} & UCF & 55.9 & 70.8 & 76.9 & 82.5 & 26.1 & 45.8 & 57.9 & 69.7 \\
 \hline
 \textbf{STC-mix} & UCF & \textbf{58.1} & \textbf{76.5} & \textbf{83.4} & \textbf{88.7} & \textbf{27.1} & \textbf{50.7} & \textbf{65.1} & \textbf{77.1} \\
 \hline
\end{tabular}\label{SOTA_retrieval}} 
}\vspace{-0.1in}
\end{table*}

\noindent In Table~\ref{SOTA_retrieval}, we provide the video retrieval results on UCF101 and HMDB51 for the STC-mix models trained on UCF101. This is a classical test for verifying if the pre-trained model learns semantic information while learning self-supervised representation. We test if a query instance clip and its nearest neighbours belong to the same category. Our STC-mix model outperforms all the representative baselines by a significant margin on both the datasets.
\begin{table}
\centering
\scalebox{0.9}{\begin{tabular}{c|c|cc}
\hline
\multirow{2}{*}{\textbf{Method}}  & \multirow{2}{*}{$\mathcal{M}$} & \multicolumn{2}{c}{\textbf{NTU-60}} \\ 
 & &  CS & CV \\
\hline
 LongTGAN~\cite{LongTGAN} & J  & 39.1 & 48.1 \\
 MS$^2$L~\cite{MS2L} & J  & 52.6 & - \\
 AS-CAL~\cite{ASCAL} & J  & 58.5 & 64.8 \\
 P\&C~\cite{su2020predict} & J  & 50.7 & 76.3 \\
 SeBiReNet~\cite{sebirenet} & J  & - & 79.7 \\
 \hline
 SkeletonCLR$^*$ (Baseline) & J + M   & 70.1 & 77.2 \\
 \rowcolor{Gray}
  \textbf{CMMC (Skeleton)} & J + M  & 72.5 & 79.1 \\ \hline
 2s-CrosSCLR~\cite{li2021crossclr} & J + M   & 74.5 & 82.1 \\
 \rowcolor{Gray}
  \textbf{CMMC (2s-Skeleton)} & J + M  & \textbf{75.2} & \textbf{83.1} \\
  \hline
\end{tabular}}
\caption{Unsupervised results on NTU-60 for action classification using different modalities $\mathcal{M}$ (J and M). $^*$ indicates that results reproduced on our settings.}\label{SOTA_ntu}
\end{table}

\noindent \textbf{Generalizing CMMC on other modalities.} In Table~\ref{SOTA_ntu}, we present SSL with pose modality for skeleton action recognition results on NTU-60. We perform CMMC with hidden skeleton representations in the encoder. We treat Joints (J) and Motion (M) as different input modalities for given skeleton data. Note that the cutmix operation across spatial dimension represents the joint vertices (1-dimensional). The downstream action classification results of a skeleton model pre-trained with contrastive learning (SkeletonCLR) using CMMC outperforms its baseline by $2.4$\% on CS and by $1$\% on CV protocol. We further improve this representation with CrosSCLR~\cite{li2021crossclr} which with its cross-modal positive mining benefits the vanilla SkeletonCLR model.  We believe that such positive mining approaches in contrastive learning such as CoCLR or CrosSCLR can benefit from our video mixing strategies. 
\begin{table}
\scalebox{0.8}{
 \begin{tabular}{c|c|c|c|c|cc}
\hline
& \multirow{2}{*}{\textbf{Method}}  & \multirow{2}{*}{\textbf{$\mathcal{M}$}} & \textbf{\small{Extra}} & \textbf{Pre-train}  & \multicolumn{2}{c}{\textbf{NTU-60}}  \\
\cline{5-7}
& & & \textbf{\small{Data}} & \textbf{Dataset} &  CS & CV  \\
\hline
{\multirow{4}{*}{\rotatebox{90}{Supervised}}} & I3D~\cite{i3d} & R & \checkmark & K400 &  85.5 & 87.3 \\
& NPL~\cite{NPL_2021_CVPR} & R & \checkmark & K400  & - & 93.7  \\
& STA~\cite{STA_iccv} & R+P & \checkmark & K400  & 92.2 & 94.6 \\
& VPN~\cite{VPN} & R+P & \checkmark & K400 & \textbf{93.5} & \textbf{96.2}  \\
\hline
{\rotatebox{90}{L}} & MoCo (S3D) & R & $\times$ & NTU-60  & 87.5 & 91.3  \\
\rowcolor{Gray}
\rotatebox{90}{S}& \textbf{STC-mix} & R & $\times$ & NTU-60 & 88.1 & 92.0 \\
\rowcolor{Gray}
\rotatebox{90}{S}& \textbf{STC-mix} & R+P & $\times$ & NTU-60 & 91.4 & 95.1 \\
\hline
\end{tabular}}
\caption{Comparison to the SOTA methods on NTU-60 using RGB (R) and Pose (P) modalities. Here, K400 indicates Kinetics~\cite{kinetics}, $\mathcal{M}$ indicates modality, and SSL indicates self-supervised learning.}\label{SOTA_ntu60}
\end{table}

\noindent \textbf{Generalizing CMMC on heterogeneous input modalities.} In an attempt to generalize STC-mix further, we perform STC-mix with RGB (S3D backbone) and Poses (ST-GCN backbone) as the cross-modal inputs. For brevity, we perform CMMC only at block 4 of S3D and ST-GCN. Moreover, we do not perform cutmix in the spatial dimension, instead we reshape the entire spatial dimension (25 joints) into a $5 \times 5$ region while mixing the skeleton features with RGB. Conversely, we always extract a $3 \times 3$ RGB region followed by vectorizing it to mix with the skeleton features. More details for this experiment is provided in the Appendix. In Table~\ref{SOTA_ntu60}, we show that STC-mix improves the baseline MoCo model. Furthermore, `self-supervised' STC-mix pre-trained with both the modalities performs on par with the SOTA `supervised' methods which are pre-trained on ImageNet~\cite{imagenet_cvpr09} and Kinetics~\cite{kinetics}.

\begin{table} 
\centering
\scalebox{0.67}{
\begin{tabular}{c|cccc|c|cc}
\hline
\textbf{Method}     & \textbf{Network} & \textbf{Res.} & $\mathcal{D}$ & $\mathcal{M}$ &  \textbf{\small{GFLOPs}}  & \textbf{UCF} & \textbf{HMDB} \\ \hline
CoCLR~\cite{coclr}& $2\times$S3D & 128 & 23  & V+F& 22  &  90.6 & 62.9 \\
\rowcolor{Gray}
\textbf{STC-mix}   & $2\times$S3D & 128 & 23  & V+F & 22 & 91.1 & 62.3 \\
\rowcolor{Gray}
\textbf{STC-mix+CoCLR}   & S3D & 128 & 23 & V  & 11 & 89.4 & 55.3 \\
\rowcolor{Gray}
\textbf{STC-mix+CoCLR}   & $2\times$S3D & 128 & 23  & V+F & 22 & 92.3 & 63.5 \\
\hline
XDC~\cite{alwassel_2020_xdc}   & R(2+1)D & 224 & 26 & V+A & 48.5  & 84.2 & 47.1 \\
AVTS~\cite{cooperative_learning} & I3D & 224 & 22 & V+A  & 108  & 83.7 & 53.0 \\
MemDPC~\cite{dpc} & $2\times$R-2D3D & 224 & 33  & V+F & 140  &  86.1 & 54.5 \\
CVRL~\cite{cvrl}  & R3D & 224 & 49 & V & 167  & 92.1 & 65.4 \\
$\rho$BYOL~\cite{video_byol}  & S3D & 224 & 23  & V & 36  & 96.3 & 75.0 \\
\rowcolor{Gray}
\textbf{STC-mix$\dagger$}   & $2\times$S3D & 224 & 23  & V+F & 72 & \textbf{96.9} & \textbf{75.4} \\
\rowcolor{Gray}
\textbf{STC-mix$\dagger$}   & S3D & 224 & 23  & V & 36 & 96.6 & 75.2 \\
\hline
\end{tabular}} \vspace{-0.1in}
\caption{Fine-tune results of models trained on Kinetics-400 for different combination of modalities - Video (V), Flow (F), and Audio (A). \textbf{STC-mix$\dagger$} is adapted with $\rho$BYOL settings.}\label{SOTA_finetune}
\end{table}

\noindent \textbf{SSL with STC-mix on large scale data.} 
In Table~\ref{SOTA_finetune}, our STC-mix model with lower FLOPs and lower input resolution (128), pre-trained on Kinetics-400 performs on par with the other SOTA models substantiating the impact of simple cross-modal video augmentation. We adapt STC-mix with $\rho$BYOL training settings replacing of MoCo type SSL for this large-scale pre-training on Kinetics. More details on this adaptation is provided in the Appendix. We observe that this variant STC-mix$\dagger$ outperforms all the SOTA SSL methods showing its usefulness in the SSL paradigm.   


\section{Related work}
Deep neural networks, especially the networks fabricated for processing videos are data-hungry. While annotating large scale video data is expensive, recently many self-supervised video representation learning approaches have been proposed to make use of the abundant web videos. On one hand, some methods have exploited the temporal structure of the videos, such as predicting if frames appear in order, reverse order, shuffled, color-consistency across frames, etc~\cite{OPN,sst2,sst3,shufflelearn,ss1,ss2,ss3,brave}. On the other hand, some methods have been taking  advantage of the multiple modalities of videos like audio, text, optical flow, etc by designing pretext tasks for their temporal alignment~\cite{out_of_time,cooperative_learning,objects_that_sounds,owens2018audiovisual,evolving_losses,miech20endtoend,look_listen}. 

Meanwhile, data mixing strategies have gained popularity in image-domain data augmentations for supervised learning~\cite{zhang2018mixup,cutoff,yun2019cutmix} in addition to their usage also for learning self-supervised image representation~\cite{manifold_mixup,lee2021imix,verma2021dacl}. A recent work (unpublished), in the spirit of data mixing in the video domain, VideoMix creates a new training video by inserting a video cuboid into another video in the supervised setting~\cite{yun2020videomix}. In contrast, we focus on mixing video samples for self-supervised representation. Different from the observations in VideoMix, we note that mixup in STC-mix is a better augmentation tool rather than strategies involving removal of spatio-temporal sub-space from the original videos. The most closest to our work, Manifold mixup~\cite{manifold_mixup} focuses on interpolating hidden representation of the samples within a mini-batch, whereas, our proposed CMMC in STC-mix performs cutmix operation in the data manifold across different modalities. In addition, we also introduce the notion of channel mixing in the feature space. We find that STC-mix is simple to implement while is a strong data augmentation tool for learning self-supervised video representation even with small data size.

\section{Conclusion}
We have analyzed the augmentation strategies for learning self-supervised video representation. We have introduced STC-mix which involves performing video mixup followed by cross-modal manifold mixup to take advantage of different modalities present in videos. STC-mix improves the quality of learned representation and thus brings large improvement in the performance of downstream tasks on UCF101, HMDB51 and NTU-60 datasets. We believe that STC-mix can be a standard video augmentation tool while learning any multi-modal self-supervised video representation. To facilitate future research, we will release our code and pre-trained representations.
 
\noindent\textbf{Acknowledgements.} 
This work was supported by the National Science Foundation (IIS-2104404). 
We thank Saarthak Kapse and Xiang Li for the great help in preparing the manuscript. 
We also thank members in the Robotics Lab at SBU for valuable discussions.

{\small
\bibliographystyle{ieee_fullname}
\bibliography{main}

\begin{thebibliography}{10}\itemsep=-1pt

\bibitem{alwassel_2020_xdc}
Humam Alwassel, Dhruv Mahajan, Bruno Korbar, Lorenzo Torresani, Bernard Ghanem,
  and Du Tran.
\newblock Self-supervised learning by cross-modal audio-video clustering.
\newblock In {\em Advances in Neural Information Processing Systems (NeurIPS)},
  2020.

\bibitem{look_listen}
Relja Arandjelovic and Andrew Zisserman.
\newblock Look, listen and learn.
\newblock {\em CoRR}, abs/1705.08168, 2017.

\bibitem{objects_that_sounds}
Relja Arandjelovic and Andrew Zisserman.
\newblock Objects that sound.
\newblock {\em CoRR}, abs/1712.06651, 2017.

\bibitem{speednet}
Sagie Benaim, Ariel Ephrat, Oran Lang, Inbar Mosseri, William~T. Freeman,
  Michael Rubinstein, Michal Irani, and Tali Dekel.
\newblock Speednet: Learning the speediness in videos.
\newblock In {\em IEEE/CVF Conference on Computer Vision and Pattern
  Recognition (CVPR)}, June 2020.

\bibitem{buchler}
Uta B{\"{u}}chler, Biagio Brattoli, and Bj{\"{o}}rn Ommer.
\newblock Improving spatiotemporal self-supervision by deep reinforcement
  learning.
\newblock {\em CoRR}, abs/1807.11293, 2018.

\bibitem{i3d}
Joao Carreira and Andrew Zisserman.
\newblock Quo vadis, action recognition? a new model and the kinetics dataset.
\newblock In {\em 2017 IEEE Conference on Computer Vision and Pattern
  Recognition (CVPR)}, pages 4724--4733. IEEE, 2017.

\bibitem{simclr}
Ting Chen, Simon Kornblith, Mohammad Norouzi, and Geoffrey Hinton.
\newblock A simple framework for contrastive learning of visual
  representations.
\newblock {\em arXiv preprint arXiv:2002.05709}, 2020.

\bibitem{out_of_time}
Joon~Son Chung and Andrew Zisserman.
\newblock Out of time: automated lip sync in the wild.
\newblock In {\em Workshop on Multi-view Lip-reading, ACCV}, 2016.

\bibitem{mars}
Nieves Crasto, Philippe Weinzaepfel, Karteek Alahari, and Cordelia Schmid.
\newblock {MARS: Motion-Augmented RGB Stream for Action Recognition}.
\newblock In {\em CVPR}, 2019.

\bibitem{STA_iccv}
Srijan Das, Rui Dai, Michal Koperski, Luca Minciullo, Lorenzo Garattoni,
  Francois Bremond, and Gianpiero Francesca.
\newblock Toyota smarthome: Real-world activities of daily living.
\newblock In {\em ICCV}, 2019.

\bibitem{VPN}
Srijan Das, Saurav Sharma, Rui Dai, Francois Bremond, and Monique Thonnat.
\newblock Vpn: Learning video-pose embedding for activities of daily living,
  2020.

\bibitem{imagenet_cvpr09}
Jia Deng, Wei Dong, Richard Socher, Li-Jia Li, Kai Li, and Li Fei-Fei.
\newblock {ImageNet: A Large-Scale Hierarchical Image Database}.
\newblock In {\em CVPR09}, 2009.

\bibitem{video_byol}
Christoph Feichtenhofer, Haoqi Fan, Bo Xiong, Ross Girshick, and Kaiming He.
\newblock A large-scale study on unsupervised spatiotemporal representation
  learning.
\newblock In {\em 2021 IEEE/CVF Conference on Computer Vision and Pattern
  Recognition (CVPR)}, pages 3298--3308, 2021.

\bibitem{sst2}
Basura Fernando, Hakan Bilen, Efstratios Gavves, and Stephen Gould.
\newblock Self-supervised video representation learning with odd-one-out
  networks.
\newblock In {\em IEEE Conference on Computer Vision and Pattern Recognition},
  2017.

\bibitem{garcia_cross_modal_distillation}
Nuno~C. Garcia, Pietro Morerio, and Vittorio Murino.
\newblock Modality distillation with multiple stream networks for action
  recognition.
\newblock In Vittorio Ferrari, Martial Hebert, Cristian Sminchisescu, and Yair
  Weiss, editors, {\em Computer Vision -- ECCV 2018}, pages 106--121, Cham,
  2018. Springer International Publishing.

\bibitem{infonce}
Michael Gutmann and Aapo Hyvärinen.
\newblock Noise-contrastive estimation: A new estimation principle for
  unnormalized statistical models.
\newblock In Yee~Whye Teh and Mike Titterington, editors, {\em Proceedings of
  the Thirteenth International Conference on Artificial Intelligence and
  Statistics}, volume~9 of {\em Proceedings of Machine Learning Research},
  pages 297--304, Chia Laguna Resort, Sardinia, Italy, 13--15 May 2010. PMLR.

\bibitem{dpc}
Tengda Han, Weidi Xie, and Andrew Zisserman.
\newblock Memory-augmented dense predictive coding for video representation
  learning.
\newblock In {\em European Conference on Computer Vision}, 2020.

\bibitem{coclr}
Tengda Han, Weidi Xie, and Andrew Zisserman.
\newblock Self-supervised co-training for video representation learning.
\newblock In {\em Neurips}, 2020.

\bibitem{he2019moco}
Kaiming He, Haoqi Fan, Yuxin Wu, Saining Xie, and Ross Girshick.
\newblock Momentum contrast for unsupervised visual representation learning.
\newblock {\em arXiv preprint arXiv:1911.05722}, 2019.

\bibitem{kinetics}
Will Kay, Joao Carreira, Karen Simonyan, Brian Zhang, Chloe Hillier, Sudheendra
  Vijayanarasimhan, Fabio Viola, Tim Green, Trevor Back, Paul Natsev, et~al.
\newblock The kinetics human action video dataset.
\newblock {\em arXiv preprint arXiv:1705.06950}, 2017.

\bibitem{cooperative_learning}
Bruno Korbar, Du Tran, and Lorenzo Torresani.
\newblock Co-training of audio and video representations from self-supervised
  temporal synchronization.
\newblock {\em CoRR}, abs/1807.00230, 2018.

\bibitem{kuehne2011hmdb}
Hildegard Kuehne, Hueihan Jhuang, Est{\'\i}baliz Garrote, Tomaso Poggio, and
  Thomas Serre.
\newblock Hmdb: a large video database for human motion recognition.
\newblock In {\em 2011 International Conference on Computer Vision}, pages
  2556--2563. IEEE, 2011.

\bibitem{OPN}
Hsin-Ying Lee, Jia-Bin Huang, Maneesh Singh, and Ming-Hsuan Yang.
\newblock Unsupervised representation learning by sorting sequences.
\newblock In {\em 2017 IEEE International Conference on Computer Vision
  (ICCV)}, pages 667--676, 2017.

\bibitem{lee2021imix}
Kibok Lee, Yian Zhu, Kihyuk Sohn, Chun-Liang Li, Jinwoo Shin, and Honglak Lee.
\newblock i-mix: A domain-agnostic strategy for contrastive representation
  learning.
\newblock In {\em ICLR}, 2021.

\bibitem{MS2L}
Lilang Lin, Sijie Song, Wenhan Yang, and Jiaying Liu.
\newblock Ms2l: Multi-task self-supervised learning for skeleton based action
  recognition.
\newblock In {\em Proceedings of the 28th ACM International Conference on
  Multimedia}, MM '20, page 2490–2498, New York, NY, USA, 2020. Association
  for Computing Machinery.

\bibitem{li2021crossclr}
Li Linguo, Wang Minsi, Ni Bingbing, Wang Hang, Yang Jiancheng, and Zhang
  Wenjun.
\newblock 3d human action representation learning via cross-view consistency
  pursuit.
\newblock In {\em CVPR}, 2021.

\bibitem{vcp}
Dezhao Luo, Chang Liu, Yu Zhou, Dongbao Yang, Can Ma, Qixiang Ye, and Weiping
  Wang.
\newblock Video cloze procedure for self-supervised spatio-temporal learning.
\newblock {\em arXiv preprint arXiv:2001.00294}, 2020.

\bibitem{miech20endtoend}
Antoine Miech, Jean-Baptiste Alayrac, Lucas Smaira, Ivan Laptev, Josef Sivic,
  and Andrew Zisserman.
\newblock {E}nd-to-{E}nd {L}earning of {V}isual {R}epresentations from
  {U}ncurated {I}nstructional {V}ideos.
\newblock In {\em CVPR}, 2020.

\bibitem{shufflelearn}
Ishan Misra, C.~Lawrence Zitnick, and Martial Hebert.
\newblock {Shuffle and Learn: Unsupervised Learning using Temporal Order
  Verification}.
\newblock In {\em ECCV}, 2016.

\bibitem{sebirenet}
Qiang Nie, Ziwei Liu, and Yunhui Liu.
\newblock Unsupervised 3d human pose representation with viewpoint and pose
  disentanglement.
\newblock In {\em European Conference on Computer Vision (ECCV)}, 2020.

\bibitem{jigsaw}
Mehdi Noroozi and Paolo Favaro.
\newblock Unsupervised learning of visual representations by solving jigsaw
  puzzles.
\newblock In {\em ECCV}, 2016.

\bibitem{owens2018audiovisual}
Andrew Owens and Alexei~A. Efros.
\newblock Audio-visual scene analysis with self-supervised multisensory
  features, 2018.

\bibitem{sst3}
Lyndsey~C. Pickup, Zheng Pan, Donglai Wei, YiChang Shih, Changshui Zhang,
  Andrew Zisserman, Bernhard Scholkopf, and William~T. Freeman.
\newblock Seeing the arrow of time.
\newblock In {\em 2014 IEEE Conference on Computer Vision and Pattern
  Recognition}, pages 2043--2050, 2014.

\bibitem{evolving_losses}
AJ Piergiovanni, Anelia Angelova, and Michael~S. Ryoo.
\newblock Evolving losses for unsupervised video representation learning.
\newblock In {\em IEEE/CVF Conference on Computer Vision and Pattern
  Recognition (CVPR)}, June 2020.

\bibitem{NPL_2021_CVPR}
AJ Piergiovanni and Michael~S. Ryoo.
\newblock Recognizing actions in videos from unseen viewpoints.
\newblock In {\em Proceedings of the IEEE/CVF Conference on Computer Vision and
  Pattern Recognition (CVPR)}, pages 4124--4132, June 2021.

\bibitem{cvrl}
Rui Qian, Tianjian Meng, Boqing Gong, Ming{-}Hsuan Yang, Huisheng Wang,
  Serge~J. Belongie, and Yin Cui.
\newblock Spatiotemporal contrastive video representation learning.
\newblock {\em CoRR}, abs/2008.03800, 2020.

\bibitem{ASCAL}
Haocong Rao, Shihao Xu, Xiping Hu, Jun Cheng, and Bin Hu.
\newblock Augmented skeleton based contrastive action learning with momentum
  lstm for unsupervised action recognition.
\newblock {\em Information Sciences}, 569:90--109, 2021.

\bibitem{brave}
Adri{\`{a}} Recasens, Pauline Luc, Jean{-}Baptiste Alayrac, Luyu Wang, Florian
  Strub, Corentin Tallec, Mateusz Malinowski, Viorica Patraucean, Florent
  Altch{\'{e}}, Michal Valko, Jean{-}Bastien Grill, A{\"{a}}ron van~den Oord,
  and Andrew Zisserman.
\newblock Broaden your views for self-supervised video learning.
\newblock {\em CoRR}, abs/2103.16559, 2021.

\bibitem{NTU_RGB+D}
Amir Shahroudy, Jun Liu, Tian-Tsong Ng, and Gang Wang.
\newblock Ntu rgb+d: A large scale dataset for 3d human activity analysis.
\newblock In {\em The IEEE Conference on Computer Vision and Pattern
  Recognition (CVPR)}, June 2016.

\bibitem{cutoff}
Dinghan Shen, Mingzhi Zheng, Yelong Shen, Yanru Qu, and Weizhu Chen.
\newblock A simple but tough-to-beat data augmentation approach for natural
  language understanding and generation.
\newblock {\em arXiv preprint arXiv:2009.13818}, 2020.

\bibitem{ucf}
Khurram Soomro, Amir~Roshan Zamir, and Mubarak Shah.
\newblock {UCF101:} {A} dataset of 101 human actions classes from videos in the
  wild.
\newblock {\em CoRR}, abs/1212.0402, 2012.

\bibitem{su2020predict}
Kun Su, Xiulong Liu, and Eli Shlizerman.
\newblock Predict \& cluster: Unsupervised skeleton based action recognition.
\newblock In {\em Proceedings of the IEEE/CVF Conference on Computer Vision and
  Pattern Recognition}, pages 9631--9640, 2020.

\bibitem{TV_L1}
Javier Sánchez~Pérez, Enric Meinhardt-Llopis, and Gabriele Facciolo.
\newblock {TV-L1 Optical Flow Estimation}.
\newblock {\em {Image Processing On Line}}, 3:137--150, 2013.

\bibitem{manifold_mixup}
Vikas Verma, Alex Lamb, Christopher Beckham, Amir Najafi, Ioannis Mitliagkas,
  David Lopez-Paz, and Yoshua Bengio.
\newblock Manifold mixup: Better representations by interpolating hidden
  states.
\newblock In Kamalika Chaudhuri and Ruslan Salakhutdinov, editors, {\em
  Proceedings of the 36th International Conference on Machine Learning},
  volume~97 of {\em Proceedings of Machine Learning Research}, pages
  6438--6447, Long Beach, California, USA, 09--15 Jun 2019. PMLR.

\bibitem{verma2021dacl}
Vikas Verma, Minh-Thang Luong, Kenji Kawaguchi, Hieu Pham, and Quoc~V Le.
\newblock Towards domain-agnostic contrastive learning.
\newblock In {\em International Conference on Machine Learning (ICML)}, 2021.

\bibitem{ss3}
Carl Vondrick, Abhinav Shrivastava, Alireza Fathi, Sergio Guadarrama, and Kevin
  Murphy.
\newblock "tracking emerges by colorizing videos".
\newblock In {\em Proceedings of the European Conference on Computer Vision
  (ECCV)}, 2018.

\bibitem{ss2}
Xiaolong Wang, Kaiming He, and Abhinav Gupta.
\newblock Transitive invariance for self-supervised visual representation
  learning.
\newblock In {\em ICCV}, 2017.

\bibitem{ss1}
Xiaolong Wang, Allan Jabri, and Alexei~A. Efros.
\newblock Learning correspondence from the cycle-consistency of time.
\newblock In {\em CVPR}, 2019.

\bibitem{s3d}
Saining Xie, Chen Sun, Jonathan Huang, Zhuowen Tu, and Kevin Murphy.
\newblock Rethinking spatiotemporal feature learning: Speed-accuracy trade-offs
  in video classification.
\newblock In Vittorio Ferrari, Martial Hebert, Cristian Sminchisescu, and Yair
  Weiss, editors, {\em Computer Vision - {ECCV} 2018 - 15th European
  Conference, Munich, Germany, September 8-14, 2018, Proceedings, Part {XV}},
  volume 11219 of {\em Lecture Notes in Computer Science}, pages 318--335.
  Springer, 2018.

\bibitem{vcop}
Dejing Xu, Jun Xiao, Zhou Zhao, Jian Shao, Di Xie, and Yueting Zhuang.
\newblock Self-supervised spatiotemporal learning via video clip order
  prediction.
\newblock In {\em Computer Vision and Pattern Recognition (CVPR)}, 2019.

\bibitem{stgcn2018aaai}
Sijie Yan, Yuanjun Xiong, and Dahua Lin.
\newblock Spatial temporal graph convolutional networks for skeleton-based
  action recognition.
\newblock In {\em AAAI}, 2018.

\bibitem{yun2019cutmix}
Sangdoo Yun, Dongyoon Han, Seong~Joon Oh, Sanghyuk Chun, Junsuk Choe, and
  Youngjoon Yoo.
\newblock Cutmix: Regularization strategy to train strong classifiers with
  localizable features.
\newblock In {\em International Conference on Computer Vision (ICCV)}, 2019.

\bibitem{yun2020videomix}
Sangdoo Yun, Seong~Joon Oh, Byeongho Heo, Dongyoon Han, and Jinhyung Kim.
\newblock Videomix: Rethinking data augmentation for video classification.
\newblock {\em arXiv preprint arXiv:2012.03457}, 2020.

\bibitem{zhang2018mixup}
Hongyi Zhang, Moustapha Cisse, Yann~N. Dauphin, and David Lopez-Paz.
\newblock mixup: Beyond empirical risk minimization.
\newblock {\em International Conference on Learning Representations}, 2018.

\bibitem{LongTGAN}
Nenggan Zheng, Jun Wen, Risheng Liu, , Liangqu Long, Jianhua Dai, and Zhefeng
  Gong.
\newblock Unsupervised representation learning with long-term dynamics for
  skeleton based action recognition.
\newblock In {\em AAAI}, pages 2644--2651, 2018.

\end{thebibliography}
}

\section*{\Large Appendix}
\appendix
\noindent \textbf{1. Training/Testing specification for downstream finetuning on UCF101 and HMDB51.} At training, we apply the same data augmentation as in the pre-training stage mentioned in section 4.2., except for Gaussian blurring. The model is trained with similar optimization configuration as in the pre-training stage for 500 epochs. At inference, we perform spatially fully convolutional inference on videos by applying ten crops (center crop and 4 corners with horizontal flipping) and temporally take clips with overlapping moving windows. The final prediction is the average \textit{softmax} scores of all the clips. 

\noindent \textbf{2. History Queue in MoCo.} We adopt momentum-updated history queue as in MoCo~\cite{he2019moco} to cache a large number of visual features while learning contrastive representation. For our pre-training experiments, we use a \textit{softmax} temperature $\tau = 0.07$, and a momentum $m = 0.999$. The queue size of MoCo for pre-training experiments on UCF101 and K400 are 2048 and 16384  respectively.

\noindent \textbf{3. More Details on Experiments with Skeletons.} \\

\textit{(i) Training/Testing specification for downstream finetuning on NTU-60.} For training the pre-trained ST-GCN along with the linear classifier, we apply the same data augmentation as in the pre-training stage. We train for 100 epochs with learning rate 0.1 (multiplied by 0.1 at epoch 80).

\textit{(ii) 2s-CrossCLR + CMMC.} For this experiment presented in Table 6., we first train each view, i.e., Joint and Motion, for 100 epochs. Then, we train the model with CMMC for another 100 epochs, and finally, we train the model with cross-view training proposed in~\cite{li2021crossclr} for 100 epochs. Thus, the model is trained for 300 epochs in total for a fair comparison with the representative baselines 2s-CrosSCLR~\cite{li2021crossclr} and SkeletonCLR.

\textit{(iii) STC-mix on NTU-60.} In this experiment, we perform STC-mix to perform cross-modal data augmentation between RGB and 3D Poses. Note that the video encoder (S3D) and skeleton encoder (ST-GCN) are asymmetric. At first, both the encoders are trained with infoNCE loss for 300 epochs. The queue size of MoCo for pre-training is set to 2048.
For brevity, we perform CMMC at fixed layers instead of randomizing the layers to be mixed across modalities. For S3D, we perform CMMC at block 4 while block 8 for ST-GCN. For training video encoder in the CMMC stage, the output of the ST-GCN from block 8 undergoes a cutmix operation with the output of block 4 of S3D. The output feature of ST-GCN is 3-dimensional ($c_2 \times t_2 \times J$), where $J=25$. Thus, we reshape $J$ into a $5 \times 5$ matrix. Thus, the resultant 4D tensor is now mixed with the video feature after undergoing a cut operation across time and channels. Conversely, while training the skeleton encoder, the intermediate RGB feature map from the 4th block of S3D is 4-dimensional ($c_2 \times t_2 \times s_2$). Thus, we perform a cut operation across all the dimensions, extracting a $3 \times 3$ crop in the spatial dimension. We flatten the spatial dimension resulting into a 3D tensor which is mixed with the ST-GCN intermediate feature. Note that this is an attempt to generalize STC-mix with asymmetric modalities. Thus, this experiment can be generalized by randomly choosing the blocks and randomizing the permissible dimension of the crops in the Cutmix operation, we leave this for future exploration. 

\begin{figure}
\begin{center}
   \includegraphics[width=1\linewidth]{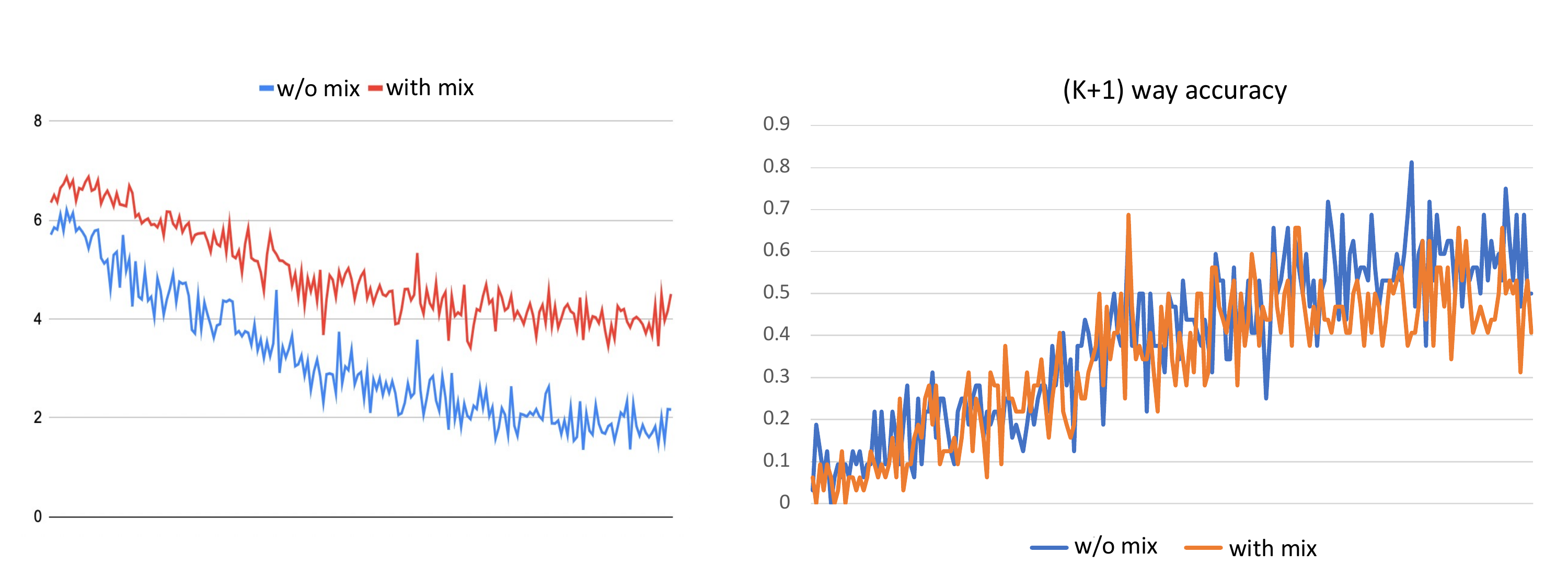} 
\end{center} 
   \caption{Training on UCF101 dataset. At left, we present the training loss of two models, one using video augmentation: STC-mix and the other not. At right, we provide the (K+1)-way accuracy on the pretext task while learning contrastive representation.}
   \label{losses}
\end{figure}
\noindent \textbf{4. Regularization effect of STC-mix}\\
In Fig.~\ref{losses}, we provide (1) a plot of training loss of two models, one using STC-mix and the other model is using standard data augmentations, and (2) the (K+1)-way accuracy of the pretext task of the models learning contrastive representation. We observe a disparity between the training losses (at left of the figure) in both the models with and without using STC-mix. This is owing to the hardness of the pretext task which can be directly correlated with the difficulty of the data transformation, via STC-mix data augmentation. Meanwhile, we also note that the (K+1)-way accuracy of the STC-mix model while training on contrastive loss is lower than that of the model without using STC-mix (at the right of the figure). However, the performance gain of the STC-mix model on downstream classification and retrieval tasks shows the regularizing capability of using STC-mix type data augmentation.
\end{document}